\pdfoutput=1

\documentclass[11pt]{article}

\usepackage[]{EMNLP2022}

\usepackage{times}
\usepackage{latexsym}

\usepackage[T1]{fontenc}

\usepackage[utf8]{inputenc}

\usepackage{microtype}

\usepackage{enumitem}
\usepackage{graphicx}
\usepackage{multirow}
\usepackage{amsmath}
\usepackage{amsfonts}
\usepackage{amssymb}
\usepackage{colortbl} 
\usepackage{stfloats}

\definecolor{ggray}{HTML}{E7E6E6}
 
\newcommand{\comment}[1]{\textcolor{black}{#1}} 
\title{Towards Unifying Reference Expression Generation and Comprehension}

\author{Duo Zheng\textsuperscript{1}\thanks{ \ \ Work was done when Zheng was interning at ByteDance AI Lab, Beijing, China.}, Tao Kong\textsuperscript{2}, Ya Jing\textsuperscript{2}, Jiaan Wang\textsuperscript{3}, Xiaojie Wang\textsuperscript{1} \\
\textsuperscript{1}Beijing University of Posts and Telecommunications \\
\textsuperscript{2}ByteDance AI Lab, Beijing, China \\
\textsuperscript{3}Soochow University, Suzhou, China \\
\texttt{\{zd, xjwang\}@bupt.edu.cn}\\
\texttt{\{kongtao,jingya\}@bytedance.com},  \texttt{jawang1@stu.suda.edu.cn} \\
}

\begin{document}
\maketitle

\begin{abstract}

Reference Expression Generation (REG) and Comprehension (REC) are two highly correlated tasks. Modeling REG and REC simultaneously for utilizing the relation between them is a promising way to improve both. However, the problem of distinct inputs, as well as building connections between them in a single model, brings challenges to the design and training of the joint model. To address the problems, we propose a unified model for REG and REC, named UniRef. It unifies these two tasks with the carefully-designed Image-Region-Text Fusion layer (IRTF), which fuses the image, region and text via the \textit{image cross-attention} and \textit{region cross-attention}. Additionally, IRTF could generate pseudo input regions for the REC task to enable a uniform way for sharing the identical representation space across the REC and REG. We further propose Vision-conditioned Masked Language Modeling (VMLM) and Text-Conditioned Region Prediction (TRP) to pre-train UniRef model on multi-granular corpora. The VMLM and TRP are directly related to REG and REC, respectively, but could help each other. We conduct extensive experiments on three benchmark datasets, RefCOCO, RefCOCO+ and RefCOCOg. Experimental results show that our model outperforms previous state-of-the-art methods on both REG and REC.
\end{abstract}

\section{Introduction}
Reference Expression (RE), which describes an unambiguous object in a real scene, is a significant cognitive behaviour in human society. People conceive a RE for an object and recognize a referent according to a RE in daily life, which we name Reference Expression Generation (REG) and Comprehension (REC), respectively. Both tasks have attracted surging interest \cite{AnnaRohrbach2015GroundingOT, ChaoruiDeng2018VisualGV, LichengYu2018MAttNetMA, ZhengyuanYang2019AFA, AishwaryaKamath2021MDETRM} from Natural Language Processing (NLP), Computer Vision (CV) and Human-Computer Interaction (HCI), due to their broad research prospects and actual applications.

REG and REC are the two sides to the same coin and are dependent on each other. For example, before conceiving an unambiguous description, people need to correctly locate the object according to the description in their mind.
However, there is less focus on addressing the unified modeling for both REG and REC. One line of the work lies in Bayes' modeling. \citet{JunhuaMao2015GenerationAC} first propose a method that can generate a RE grounded on an image, and which can also locate the object described by the RE via Bayes' rule. The subsequent work \cite{LichengYu2016ModelingCI, LichengYu2016AJS, RuotianLuo2017ComprehensionguidedRE, MikihiroTanaka2018GeneratingER, JungjunKim2020CoNANAC, JingyuLiu2020AttributeGuidedAF} typically follows this paradigm.
Another line of the work studies the parameter-shared model for the two tasks. \comment{\citet{9699024} propose the first parameter-shared framework PFOS. Considering the inputs for the two tasks are distinct (images and regions for REG while images and text for REC), PFOS shares the language-guide-vision module with the object-guide-context module, and the vision-guide-language module with the context-guide-object module. These modules need to handle the object and language inputs in REC and REG respectively, ignoring the modality gap between the inputs. To better share knowledge across REG and REC, we argue that it is important to coordinate the difference between their inputs for a unified modeling.}

\comment{Therefore, in this paper, we propose UniRef, a unified model for REG and REC. To alleviate the issue of distinct inputs, we design the Image-Region-Text Fusion layer (IRTF), which extends the transformer encoder layer through adding the \textit{image cross-attention} and \textit{region cross-attention}. Specifically, the image and region information is fused by the \textit{image cross-attention} and \textit{region cross-attention}, respectively. In REC, since the input region is not given, a region predictor is used to produce a region prediction as the input for the \textit{region cross-attention}. In this manner, UniRef could share the identical representation space across different tasks.
Furthermore, our UniRef is pre-trained with two objectives, Vison-conditioned Masked Language Modeling (VMLM) and Text-Conditioned Region Prediction (TRP) on corpora of different granularities ranging from object labels to object phrases, from region descriptions to RE.}

We note that the emergence of Vision-Langue Pre-training (VLP) \cite{JiasenLu2019ViLBERTPT, HaoTan2019LXMERTLC, LuoweiZhou2020UnifiedVP, FeiYu2020ERNIEViLKE, WeijieSu2020VLBERTPO, JaeminCho2021UnifyingVT, WonjaeKim2021ViLTVT, ZiruiWang2021SimVLMSV, AlecRadford2021LearningTV, ZhichengHuang2021SeeingOO} has greatly promoted the development of multimodal tasks. And some of them~\cite{DBLP:journals/corr/abs-2004-06165, YenChunChen2020UNITERLU, YanZeng2021MultiGrainedVL} have significantly boosted the performance of REC and demonstrated tremendous generalization ability.
Most of them focus on the alignment between either images and captions, or regions and region descriptions. 
To our knowledge, there is no VLP study focusing on unified modeling for both REG and REC.

To verify the effectiveness of our UniRef, we conduct extensive experiments on three benchmark datasets, RefCOCO, RefCOCO+ \cite{LichengYu2016ModelingCI} and RefCOCOg \cite{JunhuaMao2015GenerationAC} datasets. Experimental results deliver that our UniRef outperforms previous SOTA methods on REG and REC. 
In addition, we conduct case studies to investigate the abilities learned by our model and the challenges still remained. 

Our main contributions are concluded as follows\footnote{We release the code and model at: \url{https://github.com/zd11024/UniRef}.}:
\begin{itemize}[leftmargin=*]
\setlength{\itemsep}{0pt}
\setlength{\parsep}{0pt}
\setlength{\parskip}{0pt}

\item We propose a unified model for REG and REC, named UniRef. To alleviate the issue of distinct inputs, we design the Image-Region-Text Fusion layer (IRTF), which helps the model to share knowledge across REG and REC.
\item We pre-train UniRef with two objectives, Vision-conditioned Masked Language Modeling (VMLM) and Text-Conditioned Region Prediction (TRP), to learn the abilities required by REG and REC, respectively.
\item Experimental results show that our unified model UniRef surpasses previous SOTA models on both REG and REC.
\end{itemize}

\section{Method}
We first briefly review the task definitions of REG and REC in \S~\ref{section_task_definition}. Then we introduce the architecture of our UniRef and the pre-training in \S~\ref{section_architecture} and \S~\ref{section_pretraining}, respectively.
Last, we describe the fine-tuning and inference in \S~\ref{section_finetuning}.

\subsection{Task Definitions} \label{section_task_definition}
\vspace{0.5ex}
\noindent \textbf{Reference Expression Generation.} Given an image $I$ and a region $R$ described by box coordinates, the REG model generates the corresponding RE text $T=\{t_1,\cdots,t_{L_T}\}$ with $L_T$ tokens. The conditional distribution could be formalized as:
\begin{gather}
p_{\theta_{\mathrm{G}}}(T|I,R)=\prod_{i=1}^{L_T} p (t_i|I,R,t_{1:i-1}),
\end{gather}
where $t_{1:i-1}$ is the previous generated tokens and $\theta_{\mathrm{G}}$ are parameters of the REG model.

\vspace{0.5ex}
\noindent \textbf{Reference Expression Comprehension.} The REC model predicts the region $R$ with an image $I$ and the corresponding RE text $T$ as the input, which could be denoted as $p_{\theta_{\mathrm{C}}}(R|I,T)$. $\theta_{\mathrm{C}}$ are parameters of the REC model.

\begin{figure*}[h] 
\centering 
\includegraphics[width=1.\textwidth]{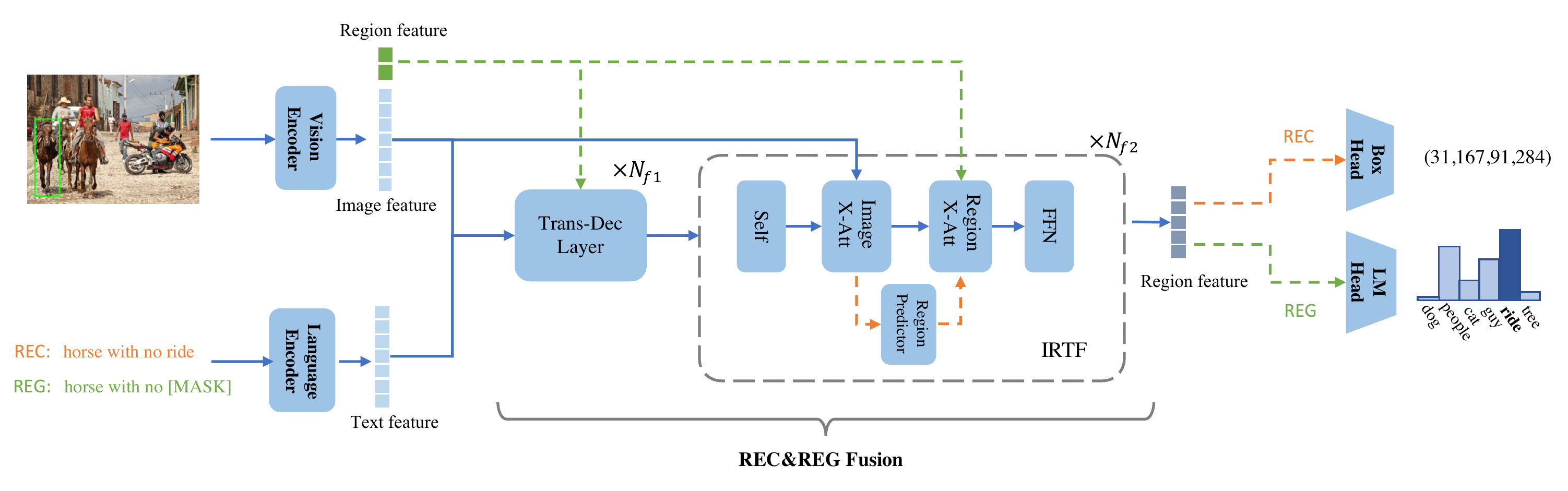} 
\caption{The architecture of UniRef. The orange and green dashed lines indicate the specific design for REC and REG respectively to enable identical representation space. ``Tran-Dec Layer'' mean the transformer decoder layer. ``Self'' means the self-attention module while ``X-Att'' represents the \textit{cross-attention} module. \comment{$N_{f1},N_{f2}$ mean the number of layers of transformer decoder block and IRTF, respectively. $\{p_i\}$ are the indexes of patches that overlap with the input bounding box.}}
\label{figure_architecture}
\end{figure*}

\subsection{Architecture} \label{section_architecture}
\comment{As depicted in Fig. \ref{figure_architecture}, UniRef consists of a vision encoder, a language encoder and a fusion encoder as well as two task-specific head, i.e., a language model (LM) head and a box head.}

\vspace{0.5ex}
    \noindent \textbf{Vision Encoder.} Given an image $I$, the vision encoder extracts the image features. It is based on the Vision Transformer (ViT) \cite{DBLP:journals/corr/abs-2004-06165} and initialized with the weights of CLIP-ViT \cite{AlecRadford2021LearningTV}. It first splits the image into non-overlapping patches, and then projects them into a sequence of embeddings. After that, these embeddings are fed to stacked transformer encoder blocks and interact with each other through self-attention, resulting in the image features $V^I=\{v_{1},\cdots, v_{L_I}\}$, where $L_I$ is the number of patches.

In REC, given a region $R$ from $I$, we obtain the region features $V^R=\{v_{p_1}, \cdots, v_{p_{L_R}}\}$, where $\{p_{i}\}$ and $L_R$ are the indexes and the number of patches that overlaps with the region, respectively.

\vspace{0.5ex}
\noindent \textbf{Language Encoder.} The language encoder is based on BERT \cite{JacobDevlin2018BERTPO}. The input sentence is tokenized into WordPieces \cite{YonghuiWu2016GooglesNM}, which are subsequently transformed into the text features $Z=\{z_{\texttt{[cls]}}, z_1, \cdots, z_{L_T}\}$ by the text encoder, where $L_T$ is the number of tokens and $z_{\texttt{[cls]}}$ \comment{are the text features corresponding to the special token \texttt{[cls]}}.

\vspace{0.5ex}
\noindent \textbf{Fusion Encoder.} 
The fusion encoder extends the transformer decoder by replacing last $N_{f2}$ vanilla transformer decoder layers with Image-Region-Text-Fusion layers (c.f., Fig. \ref{figure_architecture}), which are designed to bridge the gap between REG and REC.

The vanilla transformer decoder layer fuses region or image information via \textit{cross-attention}, depending on the input requirement of the task.

IRTF extends the vanilla transformer encoder layer through adding the \textit{image cross-attention} and \textit{region cross-attention}, and fuses the image information and region information with queries. 
Given the input $X=\{x_{\texttt{[CLS]}}, x_1, \cdots, x_{L_T}\}$, self-attention is first applied to obtain the queries:
\begin{gather}
X^{Q}=\mathrm{MHA}(X, X, X) + X,
\end{gather}
where $\mathrm{MHA}$ is multi-head attention.

Then the \textit{image cross-attention} and the \textit{region cross-attention} are performed successively as follows:
\begin{gather}
Z^{I}=\mathrm{MHA}(X^Q, V^I, V^I), \\
X^{I}=\mathrm{GLU}([Z^I, X^Q])+X^Q, \\
Z^{R}=\mathrm{MHA}(X^I, V^R, V^R), \\
X^{R}=\mathrm{GLU}([Z^R, X^Q])+X^I,
\end{gather}
where $Z^I, Z^R$ are the intermediate representations after multi-head attention. $X^I, X^R$ are the outputs of the \textit{image cross-attention} and the \textit{region cross-attention}, respectively. $[\cdot]$ means the concatenation of vectors. Following \citet{LunHuang2019AttentionOA}, we adopt Gated Linear Unit (GLU) to refine the attention outputs, denoted as:
\begin{gather}
\mathrm{GLU}(X)=\sigma(XW^1)\odot XW^2,
\end{gather}
where $W^1, W^2$ are learnable parameters, $\sigma(\cdot)$ is the sigmoid function and $\odot$ means the element-wise multiplication.

Lastly, $X^R$ is fed to a feed-forward network to obtain the output hidden states.

When performing REC, the region input is not available, which requires to predict the region conditioned on the image and text. To make the input of REC identical with REG, a region predictor is utilized for producing a region prediction, as the input for the \textit{region cross-attention}. In detail, for each patch, it calculates a score $\alpha_{i}$ based on $X_{cls}^I$ and the position embedding of $i$-th patch $e_i$. Then, it selects all patches whose scores exceed the threshold $\delta$, constituting the predicted region presentations $V^R$. We formalize this procedure as:
\begin{gather}
\label{eq_predictor0}
\alpha_i=\mathrm{MLP}([X^I_{cls}, e_i]), \\
V^R=\{V^I_i|\alpha_i\ge \delta\}.
\label{eq_predictor1}
\end{gather}

\vspace{0.5ex}
\noindent \textbf{LM Head\&Box Head.} 
To carry out REG, we use a LM head to predict the next token given the last hidden state of the \texttt{[MASK]} token. During performing REC, we employ a box head to regress the bounding box $b$ conditioned on the last hidden state of the \texttt{[CLS]} token.

\subsection{Pre-training} \label{section_pretraining}
\subsubsection{Pre-training Objectives}
To learn the abilities of language modeling and visual grounding, we pre-train UniRef with two objectives, Vision-conditioned Masked Language Modeling (VMLM) and Text-Conditioned Region Prediction (TRP), which are corresponding to REG and REC, respectively.

\vspace{0.5ex}
\noindent \textbf{Vision-Conditioned Masked Language Modeling.} Given an image-region-text triplet $(I,R,T)$, we follow X-VLM \cite{YanZeng2021MultiGrainedVL} to mask 25\% of tokens in text sequences. The task aims to predict the unseen tokens based on the visible text, region and image. Note that VMLM is similar to REG, but with differences of decoding order and attention masks. The loss function is defined as:
\begin{gather}
\mathcal{L}_{\textrm{VMLM}}=- \mathbb{E}_{(I,R,T)} \mathrm{log} \ p_{\theta_{\mathrm{G}}}(\hat T|I,R,\tilde T),
\end{gather}
where $\hat T$ and $\tilde T$ represent the masked and unmasked tokens, respectively.

\vspace{0.5ex}
\noindent\textbf{Text-Conditioned Region Prediction.} Given an image-text pair $(I,T)$, the goal of TRP is to predict the bounding box of the region or object described by the text. The loss is the summation of the generalized Intersection over Union (gIoU) \cite{HamidRezatofighi2019GeneralizedIO} and the $l_1$ distance:
\begin{gather}
\mathcal{L}_{\textrm{bbox}}=\mathbb{E}_{(I,T)} \ \mathcal{L}_{\textrm{gIoU}}(\hat b, b)+\Vert \hat b-b\Vert_1,
\end{gather}
where $\hat b, b$ represent the bounding boxes of the ground truth and prediction, respectively. 

In TRP, each IRTF produces a region prediction as the input for the \textit{region cross-attention}. The supervised signal comes from the patch-level binary cross-entropy between the prediction and the ground truth, formulated as:
\begin{gather}
\mathcal{L}_{\textrm{pred}}=\mathbb{E}_{(I,T)} \ \sum_{i}H(\hat m, m_i),
\end{gather}
where $\hat m, m_i$ mean the region mask of the ground truth and the region mask predicted by the $i$-th IRTF, respectively.

The final loss for TRP is summed by:
\begin{gather}
\mathcal{L}_{TRP}=\mathcal{L}_{\textrm{bbox}}+\mathcal{L}_{\textrm{pred}}.
\end{gather}

\subsubsection{Perspective from Probability}
We explain how our UniRef share the identical representation space across tasks in pre-training from a probability perspective. We factorize the objectives of VMLM and TRP as follows:

\begin{small}
\begin{gather}
p_{\theta_{\mathrm{G}}}(\hat T|I,R,\tilde T)=p_{\theta_{\mathrm{F}}}(H|I,R,\tilde T) \ p_{\theta_{\mathrm{LM}}}(\hat T|H),
\end{gather}
\end{small}

\begin{small}
\begin{align}
\begin{split}
p_{\theta_{\mathrm{C}}}(R|I,T)= 
&\ p_{\theta_{\mathrm{F}}}(H|I,R^{'},T)\ p_{\theta_{\mathrm{Box}}}(R|H)\\
&\ p_{\theta_{\mathrm{P}}}(R^{'}|I,T),
\end{split}
\end{align}
\end{small}where $\theta_{\mathrm{LM}},\theta_{\mathrm{Box}},\theta_{\mathrm{F}},\theta_{\mathrm{P}}$ mean the parameters of the LM head, box head, fusion encoder and predictor, respectively. $H$ are the last hidden states.
With the help of the predictor, both VMLM and TRP aim to align the region with text ($(R,\hat T)$ and $(R^{'}, T)$) conditioned on the image $I$.

\begin{table}[]
\renewcommand\arraystretch{1.}
\centering
\resizebox{0.45\textwidth}{!}{
\setlength{\tabcolsep}{1.2mm}{
    \begin{tabular}{c|ccc} \hline
        Pre-training Dataset & \# Images & \# Text & Avg. Tok \\ \hline
        \multirow{1}{*}{\shortstack{COCO Object Labels}} & \multirow{1}{*}{112k} & \multirow{1}{*}{434k} & \multirow{1}{*}{1.20} \\
        \multirow{1}{*}{\shortstack{VG Phrases}} & \multirow{1}{*}{104k} & \multirow{1}{*}{2M} & \multirow{1}{*}{1.24} \\
        \multirow{1}{*}{\shortstack{VG Region Descriptions}} & \multirow{1}{*}{105k} & \multirow{1}{*}{360k} & \multirow{1}{*}{5.40} \\
        \multirow{1}{*}{\shortstack{RefCOCO-MERGE}} & \multirow{1}{*}{24k} & \multirow{1}{*}{287k} & \multirow{1}{*}{5.07} \\
        \hline
    \end{tabular}
}
}
\caption{The statistics of the pre-training datasets. ``\# Images ''and ``\# Text'' represent the number of images and text descriptions, ``Avg. Tok'' indicates the average number of tokens in descriptions.} 
\vspace{-0.3cm}
\label{table_statistics}
\end{table}

\subsubsection{Pre-training Datasets} \label{secton_pre-training_data}
We collect four pre-training datasets of different granularities ranging from object labels to pharases, from region descriptions to RE: (1) COCO object labels \cite{TsungYiLin2014MicrosoftCC}. Each object corresponds to a label in 80 pre-defined categories. (2) Visual Genome (VG) phrases \cite{RanjayKrishna2017VisualGC}. We concatenate the attribute and object of an object to form a phrase. There are over 75k unique objects and 50k unique attributes, leading to more combinations of objects and attributes. (3) Visual Genome region descriptions. The region descriptions could be either a phrase or a sentence. (4) RefCOCO-MERGE. We merge RefCOCO, RefCOCO+ and RefCOCOg together. For the above datasets, we filter out the data whose image appears in the val and test set of RefCOCO, RefCOCO+, RefCOCOg according to COCO id. Tab. \ref{table_statistics} lists the statistics of the pre-training datasets.

\begin{table*}[]
\centering
\renewcommand\arraystretch{1}
\resizebox{1\textwidth}{!}{
\setlength{\tabcolsep}{2.5mm}{
\begin{tabular}{l|c|cccc|cccc|cccc} \hline
\multirow{3}{*}{Method} & \multirow{3}{*}{\shortstack{\# Pre-train\\Images}}& \multicolumn{4}{c}{RefCOCO} & \multicolumn{4}{|c|}{RefCOCO+} & \multicolumn{4}{c}{RefCOCOg} \\ \cline{3-14}
& & \multicolumn{2}{c}{testA} & \multicolumn{2}{c}{testB} & \multicolumn{2}{|c}{testA} & \multicolumn{2}{c|}{testB} & \multicolumn{2}{c}{val} & \multicolumn{2}{c}{test} \\ \cline{3-14}
& & M & C & M & C & M & C& M & C & M & C & M & C \\ \hline
SR \citeyearpar{Tanaka_2019_ICCV} & - & 0.301 & 0.866 & 0.341 & 1.389 & 0.243 & 0.672 & 0.222 & 0.831 & 0.160 & 0.741 & 0.160 & 0.727 \\
SR-rerank \citeyearpar{Tanaka_2019_ICCV} & - & 0.310 & 0.842 & 0.348 & 1.356 & 0.241 & 0.656 & 0.219 & 0.782 & 0.165 & 0.756 & 0.164 & 0.764 \\
CoNAN \citeyearpar{JungjunKim2020CoNANAC} & - & 0.330 & 0.915 & 0.354 & 1.410 & 0.288 & 0.761 & 0.250 & 0.876 & - & - & - & - \\
VL-T5 \citeyearpar{JaeminCho2021UnifyingVT} & 180k & 0.334 & 0.978 & 0.347 & 1.427 & 0.288 & 0.828 & 0.245 & 0.852 & 0.189 & 0.873 & 0.189 & 0.881 \\
UniRef & 180k & \textbf{0.347} & \textbf{1.049} & \textbf{0.374} & \textbf{1.549} & \textbf{0.311} & \textbf{0.916} & \textbf{0.266} & \textbf{0.972} & \textbf{0.197} & \textbf{1.033} & \textbf{0.195} & \textbf{1.017} \\
\hline
\end{tabular}}
}
\caption{The performance on REG. ``M'' and ``C'' indicate Meteor and CIDEr, respectively. ``-'' means that the details are not reported. ``\# Pre-train Images'' means the number of images in pre-training datasets. }
\label{table_performace_reg}
\end{table*}

\begin{table*}[]
\centering
\renewcommand\arraystretch{1}
\setlength{\tabcolsep}{2.5mm}{
\begin{tabular}{l|cc|cc|cc|cc} \hline
\multirow{2}{*}{Method} &\multirow{2}{*}{\shortstack{\# Params}} & \multirow{2}{*}{\shortstack{\# Pre-train\\Images}} & \multicolumn{2}{c}{RefCOCO} & \multicolumn{2}{|c|}{RefCOCO+} & \multicolumn{2}{c}{RefCOCOg} \\ \cline{4-9}
& & & testA & testB & testA & testB & val & test \\ \hline
MattNet \citeyearpar{LichengYu2018MAttNetMA} & - & - & 81.14 & 69.99 & 71.62 & 56.02 & 66.58 & 67.27 \\
ViLBERT \citeyearpar{JiasenLu2019ViLBERTPT} & - & 3.3M & - & - & 78.52 & 62.61 & - & - \\
VL-BERT\textsubscript{large} \citeyearpar{WeijieSu2020VLBERTPO} & - & 3.3M & - & - & 78.57 & 62.30 & - & - \\
UNITER\textsubscript{large} \citeyearpar{YenChunChen2020UNITERLU} & 300M  & 3.3M & - & - & 78.57 & 62.30 & - & - \\
MDETR \citeyearpar{AishwaryaKamath2021MDETRM} & - & 200k & 90.42 & 83.06 & 85.05 & 71.88 & 83.44 & 83.93 \\
X-VLM \citeyearpar{YanZeng2021MultiGrainedVL} & 240M & 4M & - & - & 86.36 & 71.00 & - & - \\
OFA \citeyearpar{PengWang2022UNIFYINGAT} & 180M & 14.7M & 90.67 & 83.30 & 87.15 & 74.29 & 82.29 & 82.31 \\
UniRef & 227M & 180k & \textbf{91.21} & \textbf{83.87} & \textbf{87.74} & \textbf{75.45} & \textbf{85.62} & \textbf{84.92} \\ 
\hline
\end{tabular}}
\caption{The accuracy (\%) on REC. ``-'' means that the details are not reported. ``\# Pre-train Images'' means the number of images in pre-training datasets. \comment{The comparing models are base-size unless otherwise specified.}}
\label{tabel_performance_rec}
\end{table*}

\subsection{Fine-tuning and Inference.} \label{section_finetuning}
Following \citet{DBLP:journals/corr/abs-2004-06165}, we fine-tune UniRef for REG on RefCOCO, RefCOCO+ and RefCOCOg separately. In detail, 25\% of the tokens are randomly masked and the model recovers them with a unidirectional attention mask instead of a bidirectional one. During inference, at each step, a \texttt{[MASK]} token is appended to the end of current generation, with a subsequent forward-pass to generate the next token. The process terminates until a \texttt{[SEP]} token is produced.
For REC, the procedure is same to TRP. 

\section{Experiments}
\subsection{Datasets and Metrics}
\noindent \textbf{Datasets.} We evaluate our model on three widely-used benchmark datasets, i.e., RefCOCO, RefCOCO+ \cite{LichengYu2016ModelingCI} and RefCOCOg \cite{JunhuaMao2015GenerationAC}, which are based on COCO \cite{TsungYiLin2014MicrosoftCC} images. 

RefCOCO contains 142,209 reference expressions for 50,000 objects on 19,994 images, while RefCOCO+ consists of 141,564 descriptions for 50,000 objects on 19,992 images. 
Their test sets are split into testA and testB by ``People vs. Object''.
The main difference is that position words are prohibited in RefCOCO+, leading to more appearance-centric descriptions.

ReCOCOg contains 54,822 objects on 26,711 images with 104,560 expressions, which are longer and more informative than that of RefCOCO/RefCOCO+.
For RefCOCOg, most methods evaluate on Google split in REG, and on UMD split in REC. In this paper, we reproduce some representative REG methods on UMD split and report the corresponding results.

\vspace{0.5ex}
\noindent \textbf{Metrics.} We evaluate the performance of REG with two automatic metrics, i.e., CIDEr \cite{RamakrishnaVedantam2014CIDErCI} and Meteor \cite{AlonLavie2009TheMM}. In REC, we report the accuracy of bounding box prediction. A prediction is correct if its IoU with the ground truth is greater than 0.5.

\begin{table*}[]
\centering
\renewcommand\arraystretch{1}
\resizebox{1\textwidth}{!}{
\setlength{\tabcolsep}{1.mm}{
\begin{tabular}{l|l|ccccccc|ccccccc} \hline
\multirow{3}{*}{\#} &  & \multicolumn{7}{c|}{\textbf{REG}} & \multicolumn{7}{c}{\textbf{REC}} \\
& & \multicolumn{1}{c}{\cellcolor{ggray!90}{}} & \multicolumn{2}{c}{RefCOCO} & \multicolumn{2}{c}{RefCOCO+} & \multicolumn{2}{c|}{RefCOCOg} & \multicolumn{1}{c}{\cellcolor{ggray!90}{}} & \multicolumn{2}{c}{RefCOCO} & \multicolumn{2}{c}{RefCOCO+} & \multicolumn{2}{c}{RefCOCOg} \\
& & \multicolumn{1}{c}{\cellcolor{ggray!90}{\multirow{-2}{*}{Avg.}}} & testA & testB & testA & testB & val & test & \multicolumn{1}{c}{\cellcolor{ggray!90}{\multirow{-2}{*}{Avg.}}} & testA & testB & testA & testB & val & test \\
\hline 
1& UniRef (no IRTF) & \cellcolor{ggray!90}{1.075} & 1.041 & 1.513 & 0.895 & 0.977 & 1.011 & 1.010 & \cellcolor{ggray!90}{83.99} & 90.29 & 83.74 & 86.38 & 75.55 & 83.84 & 84.11 \\ 
2& UniRef (IRTF in L4,5,6) & \cellcolor{ggray!90}{1.088} & 1.063 & 1.540 & 0.910 & 0.988 & 1.015 & 1.012 & \cellcolor{ggray!90}{84.44} & 90.79 & 84.07 & 86.74 & 74.45 & 85.27 & 85.30 \\
3 & UniRef (IRTF in L5,6) & \cellcolor{ggray!90}{1.083} & 1.031 & 1.505 & 0.912 & 0.981 & 1.037 & 1.033 & \cellcolor{ggray!90}{\underline{84.68}} & 91.04 & 84.44 & 87.08 & 75.53 & 85.01 & 84.95 \\ \hline\hline

4 &UniRef (IRTF in L6) & \cellcolor{ggray!90}{\underline{1.089}} & 1.049 & 1.549 & 0.916 & 0.972 & 1.033 & 1.017 & \cellcolor{ggray!90}{\textbf{84.80}} & 91.21 & 83.87 & 87.74 & 75.45 & 85.62 & 84.92 \\
5 &\quad \textit{w/o.} GLU & \cellcolor{ggray!90}{1.080} &  1.054 & 1.511 & 0.899 & 0.985 & 1.015 & 1.014 & \cellcolor{ggray!90}{84.65} & 90.93 & 84.81 & 86.78 & 75.72 & 85.44 & 84.20 \\
6&\quad \textit{w/o.} VMLM & \cellcolor{ggray!90}{0.760} & 0.818 & 1.183 & 0.645 & 0.738 & 0.591 & 0.585 & \cellcolor{ggray!90}{82.46} & 89.50 & 82.99 & 84.51 & 72.51 & 82.68 & 82.56 \\ 
7&\quad \textit{w/o.} TRP & \cellcolor{ggray!90}{1.060} & 1.025 & 1.492 & 0.889 & 0.962 & 1.003 & 0.989 & \cellcolor{ggray!90}{61.39} & 75.06 & 63.96 & 65.81 & 48.98 & 57.84 & 56.67 \\
8&\quad \textit{w/o.} RefCOCO-MERGE & \cellcolor{ggray!90}{\textbf{1.098}} & 1.063 & 1.540 & 0.910 & 0.988 & 1.043 & 1.043 & \cellcolor{ggray!90}{82.31} & 89.52 & 82.68 & 84.23 & 71.01 & 83.35 & 83.07 \\
\hline
\end{tabular}}
}
\caption{The ablation studies of fusion encoder and pre-training. We report CIDEr for REG and accuracy for REC. \colorbox{ggray!90}{Avg.} means the average of CIDEr/accuracy on REG/REC. ``UniRef (IRTF in LX)'' means that layers X are IRTF while others are transformer decoder layers, and ``UniRef (no IRTF)'' indicates the fusion encoder only contains transformer decoder layers. The \textbf{bold} and \underline{underline} denote the best and the second performances, respectively.}
\vspace{-0.2cm}
\label{table_ablation_pretraining}
\end{table*}

\subsection{Implementation Details} \label{section_implementation_details}
The vision encoder of UniRef is initialized with weights of CLIP-ViT/B-16\footnote{\url{https://huggingface.co/openai/clip-vit-base-patch16}}. The text and fusion encoder is initialized with weights of the first six and last six layers of BERT\textsubscript{base}, respectively. The extra parameters of the fusion encoder, including the \textit{cross-attention} and predictor, are randomly initialized. For the fusion encoder, we adopt vanilla transformer decoder layers as the first five layers and IRTF as the last layer.

We implement our method with Pytorch and perform all experiments on NVIDIA Tesla A100 GPU. We pre-train UniRef for 200k steps with a batch size of 1024. The learning rate is warmed-up from 1e-5 to 1e-4, with a subsequent decay to 1e-5. In the fine-tuning stage, we train REG and REC models for 20 epochs with a batch size of 40. Following \citet{YanZeng2021MultiGrainedVL}, the image resolution is set to 224 in pre-training while 384 in fine-tuning.

\subsection{Comparing Models}
\comment{In this section, we compare UniRef with the SOTA models of REG and REC, respectively.}

\noindent \textbf{REG.}
(1) SR \cite{Tanaka_2019_ICCV} extends the speaker-listener-reinforcer framework \cite{LichengYu2016AJS} with a well-designed attention mechanism. (2) SR-rerank picks the expression through reranking a set of generated sentences. 
(3) CoNAN \cite{JungjunKim2020CoNANAC} introduces an attentional ranking module to obtain complementary neighbor features.
(4) VL-T5 \cite{JaeminCho2021UnifyingVT} unifies many tasks into a sequence-to-sequence framework via instruction learning. To adapt VL-T5 to REG, we append the region features at the fixed position of the input.

\vspace{0.5ex}
\noindent \textbf{REC.}
(1) MattNet \cite{LichengYu2018MAttNetMA} is a representative two-stage method. (2) ViLBERT \cite{JiasenLu2019ViLBERTPT}, (3) VL-BERT\textsubscript{large} \cite{WeijieSu2020VLBERTPO} and (4) UNITER\textsubscript{large} \cite{YenChunChen2020UNITERLU} are VLP models with region features. (5) MDETR \cite{AishwaryaKamath2021MDETRM} is a pre-trained model that takes DETR \cite{NicolasCarion2020EndtoEndOD} as the backbone. 
Additionally, (6) X-VLM \cite{YanZeng2021MultiGrainedVL} and (7) OFA \cite{PengWang2022UNIFYINGAT} are pre-trained on much larger datasets and show marvelous generalization ability. Note that X-VLM and OFA also utilize fine-grained labeled data, thus the comparison is fair.

\subsection{Main Results}
\comment{In REG and REC, our UniRef delivers better results than previous SOTA results, which cannot be simultaneously achieved by previous methods.}
\paragraph{Performance on REG.}
As shown in Tab. \ref{table_performace_reg}, our UniRef outperforms previous SOTA methods on three datasets. Specifically, UniRef achieves 1.049/1.549 on RefCOCO testA/testB, 0.916/0.972 on RefCOCO+ testA/testB, and 1.033/1.017 on RefCOCOg val/test, in terms of CIDEr. Furthermore, it has the most prominent improvement on RefCOCOg, with CIDEr lift rate of 18.3\% and 15.4\% on val and test respectively, compared with VL-T5. This demonstrates that our model can better handle the expression with more details.
\paragraph{Performance on REC.}
As shown in Tab. \ref{tabel_performance_rec}, our UniRef outperforms SOTA models on all benchmark datasets.
Specifically, it outperforms MDETR by 0.79/0.81\% on RefCOCO, 2.69/3.57\% on RefCOCO+ and 2.18/0.99\% on RefCOCOg. 
Even compared to OFA pre-trained on 14.7M images, our model still shows its superiority, especially on RefCOCOg.

\begin{figure*}[h] 
\centering 
\includegraphics[width=0.96\textwidth]{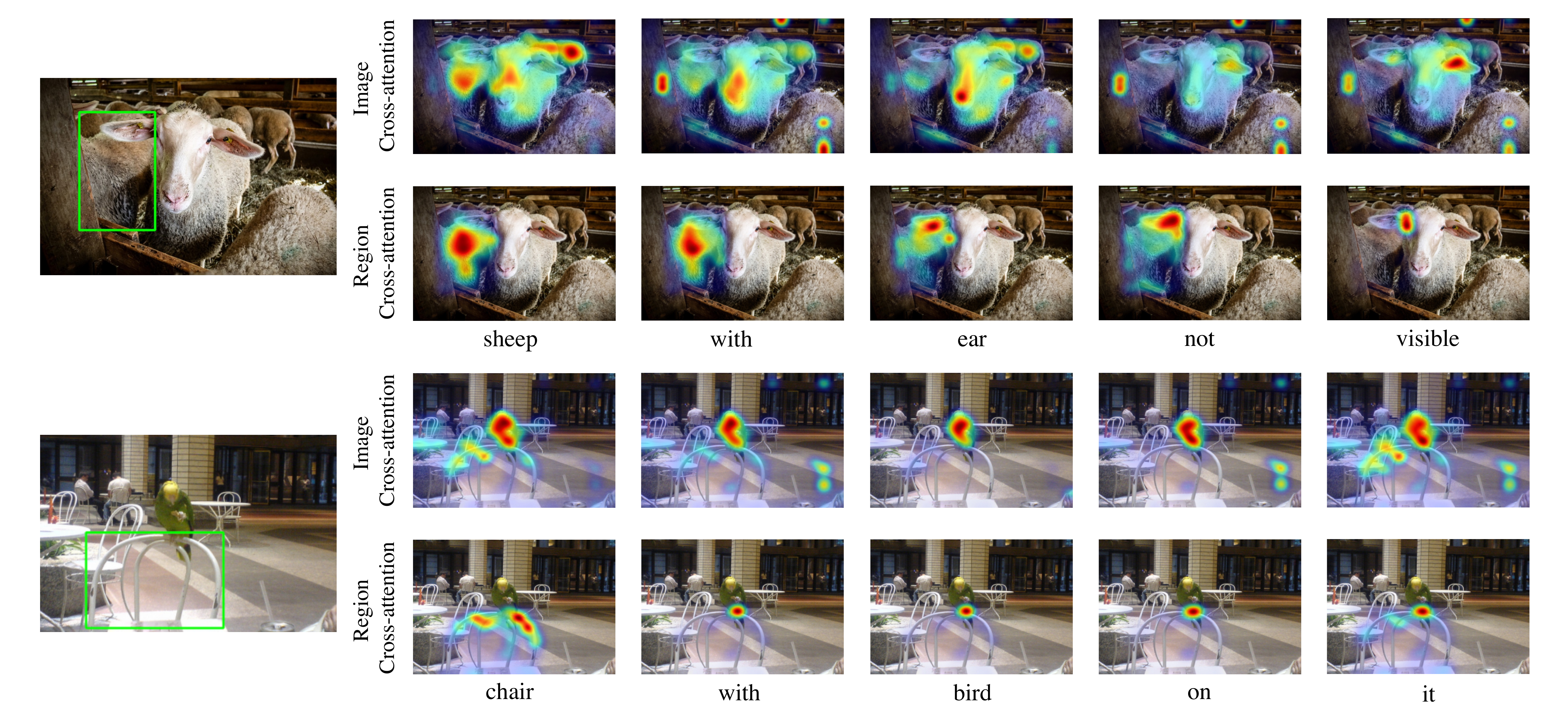} 
\vspace{-0.2cm}
\caption{The visualization of the autoregressive generation process of REG. The model successively pays attention to the whole image and region, with a subsequent generated token. For each example, the first and second row respectively give the attention maps of \textit{image cross-attention} and the \textit{region cross-attention} when generating the next token, which is below the corresponding \textit{cross-attention} maps. The input region is marked in the green box.}
\vspace{-0.3cm}
\label{figure_case_study2}
\end{figure*}

\subsection{Ablation Study.}
To investigate the effects of the fusion encoder and pre-training, we conduct ablation studies (Tab. \ref{table_ablation_pretraining}).

\vspace{0.5ex}
\noindent \textbf{ITRF Boosts the Results on REG and REC.} 
Comparing rows 1 and 4, it can be seen that the UniRef with IRTF in 6-th layer outperforms the counterpart without IRTF, validating the effectiveness of IRTF. 
IRTF decouples the \textit{cross-attention} into image and \textit{region cross-attention}, and takes image, region and text as the identical inputs, resulting in better interaction between them.
Furthermore, GLU slightly boost the performance for it could refine the attention outputs via non-linear transformation (row 4 vs. row 5).

\vspace{0.5ex}
\noindent \textbf{UniRef with IRTF in 6-th Layer Outperforms Other Counterparts.}
Comparing rows 2, 3 and 4, UniRef with IRTF in 6-th layer achieves the best performance. With the increase of the number of IRTF, REC performance shows a downward trend, possibly due to the error accumulation of predicted regions generated by IRTF.

\vspace{0.5ex}
\noindent \textbf{VMLM and TRP Benefit the Pre-training.}
Comparing rows 4, 6 and 7, our model outperforms the variant removing either pre-training task. The performance of REG/REC noticeably drops without VMLM/TRP, illustrating the effectiveness of the pre-training tasks.

\vspace{0.5ex}
\noindent \textbf{Pre-training on In-domain Data Significantly Improves REC but Slightly Damages REG.}
Furthermore, with pre-training on refCOCO-MERGE, UniRef suffers a significant increase in REC, from 82.31\% to 84.72\% on the average accuracy (row 8 vs. row 4). However, the average CIDEr slightly decreases in REG. We speculate that it is caused by the unbalanced sampling on the collected pre-training datasets, leading to overfitting to RefCOCO-MERGE.

\subsection{Case Study.}
In this section, we conduct case studies to provide a deeper understanding for UniRef. More examples are given in Appendix \ref{section::examples}.

\vspace{0.5ex}
\noindent \textbf{How UniRef Utilizes Image and Region Information in REG?}
As shown in Fig. \ref{figure_case_study2}, we give visualization on the \textit{cross-attention} maps, including \textit{image cross-attention} and \textit{region cross-attention}, across the process of autoregressive generation. Through observing cases, we discover two phenomena: 
1) The \textit{image cross-attention} could pay attention to other objects in the image that are indistinguishable from the target object, thereby assisting the model to generate more discriminative descriptions. For example, in the first instance, the ears of sheep are attended by \textit{image cross-attention} while the sheep with ear not visible is attended by the \textit{region cross-attention}, resulting in the description ``sheep with ear not visible''.
2) Through attending to the object related to the target object, the model could generate descriptions with relationships, e.g., spatial relationships. In the second example, the model unambiguously describes the chair in green box by the spatial relationship between it and the bird, which is not in green box.

\begin{figure*} 
\centering 
\includegraphics[width=1\textwidth]{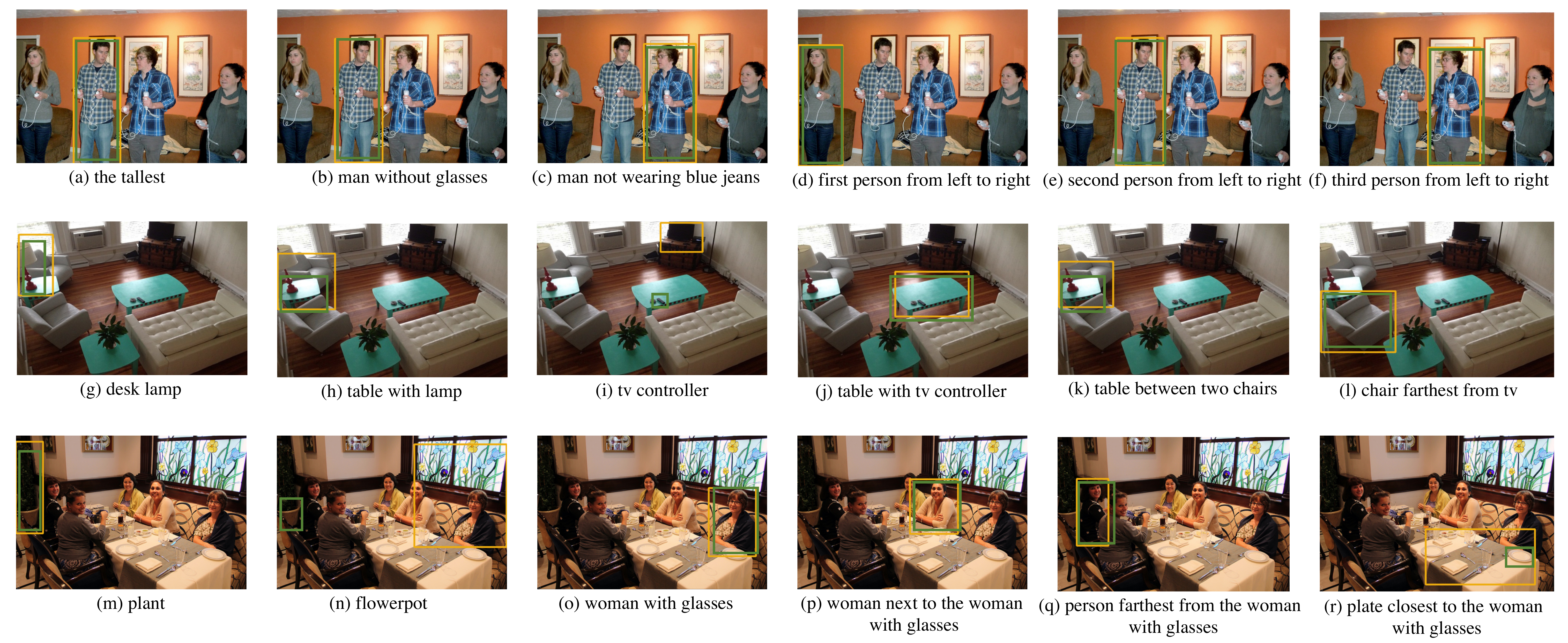} 
\vspace{-0.3cm}
\caption{Examples of the predicted bounding box in REC. The green and orange boxes indicate the ground truth and predicted boxes, respectively. The images are from RefCOCO+ while the texts are constructed.}
\vspace{-0.2cm}
\label{figure_case_study1}
\end{figure*}

\vspace{0.5ex}
\noindent \textbf{The Ability that UniRef Learns in REC.}
We give examples of bounding box predictions in Fig. \ref{figure_case_study1}. UniRef is able to handle descriptions with various properties, e.g., comparisons (Fig. \ref{figure_case_study1} (a)), attribute recognition (Fig. \ref{figure_case_study1} (b),(c)), spatial relationships (Fig. \ref{figure_case_study1} (j),(k)) and counting (Fig. \ref{figure_case_study1} (d)-(f)).

\vspace{0.5ex}
\noindent \textbf{The Challenges still Remain in REC.}
By analysing bad cases, we conclude some difficulties faced by our model: (1) Short path. The model correctly localizes the plant (Fig. \ref{figure_case_study1} (m)) while fails to ground to the flowerpot (Fig. \ref{figure_case_study1} (n)). It first locates the flowers on the wall, and then regards this wall as flowerpot. 
It shows that the model does not really understand what is flowerpot, but learns short paths through flowers; (2) Small objects. We discover that the model is not very good for small objects (Fig. \ref{figure_case_study1} (i) and (r)).

\section{Related Work}
\noindent \textbf{Reference Expression (RE).}
To study the RE, many datasets have been introduced, including RefCOCO \cite{LichengYu2016ModelingCI}, RefCOCO+ \cite{{LichengYu2016ModelingCI}} and RefCOCOg \cite{JunhuaMao2015GenerationAC}. The first two are collected in a two-palyer cooperative game, namely ReferIt \cite{SaharKazemzadeh2014ReferItGameRT}, while the last one is annotated in a non-interactive setting. 

The early work focuses on the CNN-LSTM framework, which could be applied to REG, as well as REC via Bayes' rule. 
Specifically, it first models $P(T|I,R)$, then obtains $P(R|I,T)$ by Bayes' rule, where $I,R,T$ represent the image, the reigon and the text, repectively.
\citet{JunhuaMao2015GenerationAC} first introduce this approach and propose a maximum mutual information method, which penalizes the likelihood of the RE to wrong objects in an image. 
Following this method, \citet{LichengYu2016ModelingCI} propose a visual comparative method, VisDiff, which uses the image, target object and visual difference information for generating unambiguous descriptions.
Further, \citet{LichengYu2016AJS} extend VisDiff to a speaker-listener-reinforcer model, in which the speaker, listener and reinforcer interact with each other.

Thanks to the success of object detection, REC attracts more attention and many endeavors have been devoted to it, ranging from two-stage to one-stage approaches. The two-stage methods \cite{LichengYu2018MAttNetMA, ChaoruiDeng2018VisualGV, DBLP:journals/corr/abs-1812-04794} first extract region proposals with a object detector such as faster-RCNN \cite{ShaoqingRen2015FasterRT}, then select a region conditioned on the input text. In contrast, the one-stage methods \cite{ZhengyuanYang2019AFA, ZhengyuanYang2020ImprovingOV, MuchenLi2021ReferringTA} directly predict the bounding box given the image and the text, obtaining improvement of performance from end-to-end training. 

\vspace{0.5ex}
\noindent \textbf{Vision-Language Pre-training (VLP).}
VLP, motivated by the pre-trained language models in NLP, aims at learning generic representations from abundant image-text data, advancing many vision-language tasks, e.g., VQA \cite{StanislawAntol2015VQAVQ}, image captioning and visual dialog \cite{HarmdeVries2016GuessWhatVO, AbhishekDas2017VisualD}. ViLBERT \cite{JiasenLu2019ViLBERTPT} pioneers the adaption of pre-trained models for this field. Then, VL-BERT \cite{WeijieSu2020VLBERTPO} and LXMERT \cite{HaoTan2019LXMERTLC} use a two-stream architecture for fusing information from different modality. Subsequently, \citet{DBLP:journals/corr/abs-2004-06165} propose OSCAR, which takes object labels as anchors for aligning objects and text. 
More recently, \citet{YanZeng2021MultiGrainedVL} adopt vision transformers to extract visual features and design the task of region prediction to model the fine-grained alignment between regions and descriptions.

Moreover, various technologies are applied in VLP, ranging from contrastive learning \cite{WeiLi2021UNIMOTU, AlecRadford2021LearningTV} to knowledge distillation \cite{JunnanLi2021AlignBF}, from stage-wise pre-training \cite{TongtongLiu2021MultistagePO, WenhuiWang2021VLMoUV} to prompt learning \cite{MariaTsimpoukelli2021MultimodalFL, PengWang2022UNIFYINGAT, jin-etal-2022-good}.
Standing on the shoulders of giants, we step forward with the purpose of building more advanced models for REG and REC.

\section{Conclusions}
In this paper, we propose a unified model for reference expression generation and comprehension, named UniRef. To alleviate the issue of distinct inputs for the tasks, we design the Image-Region-Text Fusion layer (IRTF) to handle the difference between the distinct inputs. In addition, UniRef is pre-trained with two objectives, Vision-conditioned Masked Language Modeling (VMLM) and Text-Conditioned Region Prediction (TRP), on multi-granular corpora. Experimental results show that our UniRef outperforms previous state-of-the-art methods on both REG and REC.

\section*{Ethical Considerations}
In this section, we consider potential ethical issues of our model.
In this paper, we propose UniRef, whose vision encoder and language encoder are initialized with the weights of CLIP-ViT \cite{AlecRadford2021LearningTV} and BERT \cite{JacobDevlin2018BERTPO}, respectively.
The pre-training datasets are collected from COCO \cite{TsungYiLin2014MicrosoftCC}, Visual Genome \cite{RanjayKrishna2017VisualGC}, RefCOCO \cite{LichengYu2016ModelingCI}, RefCOCO+ \cite{LichengYu2016ModelingCI} and RefCOCOg \cite{JunhuaMao2015GenerationAC}.
Therefore, UniRef might involve the same biases and toxic behaviors exhibited by the pre-trained models and pre-training datasets.

\section*{Limitations}
Our work has several limitations that can be further explored. (1) The size of the model and pre-training datasets could be scaled up. Since our model is designed for REG and REC, it requires carefully modification for the model architecture to adapt to massive image-text pairs. (2) We do not perform any optimization approaches for the REG model, such as self-critical sequence training and reinforcement learning. These approaches are proved to be beneficial in previous work \cite{LichengYu2016AJS, LunHuang2019AttentionOA, MarcellaCornia2019MeshedMemoryTF}. (3) It is feasible to adapt our model to other related downstream tasks, e.g., phrase grounding \cite{BryanAPlummer2015Flickr30kEC}, reference expression segmentation \cite{ChenyunWu2020PhraseCutLI} and dense captioning \cite{JustinJohnson2015DenseCapFC}, through elaborating task-specific designs. (4) It is worth more exploration on multi-task fine-tuning with REG and REC. We have done experiments that jointly fine-tune one model for both REG and REC. The performance on REG and REC is on par with or slightly worse than the separated UniRef.

\bibliography{anthology,custom}

\begin{thebibliography}{55}
\expandafter\ifx\csname natexlab\endcsname\relax\def\natexlab#1{#1}\fi

\bibitem[{Antol et~al.(2015)Antol, Agrawal, Lu, Mitchell, Batra, Zitnick, and
  Parikh}]{StanislawAntol2015VQAVQ}
Stanislaw Antol, Aishwarya Agrawal, Jiasen Lu, Margaret Mitchell, Dhruv Batra,
  C.~Lawrence Zitnick, and Devi Parikh. 2015.
\newblock \href {https://doi.org/10.1109/ICCV.2015.279} {{VQA:} visual question
  answering}.
\newblock In \emph{2015 {IEEE} International Conference on Computer Vision,
  {ICCV} 2015, Santiago, Chile, December 7-13, 2015}, pages 2425--2433. {IEEE}
  Computer Society.

\bibitem[{Carion et~al.(2020)Carion, Massa, Synnaeve, Usunier, Kirillov, and
  Zagoruyko}]{NicolasCarion2020EndtoEndOD}
Nicolas Carion, Francisco Massa, Gabriel Synnaeve, Nicolas Usunier, Alexander
  Kirillov, and Sergey Zagoruyko. 2020.
\newblock End-to-end object detection with transformers.
\newblock \emph{european conference on computer vision}.

\bibitem[{Chen et~al.(2020)Chen, Li, Yu, Kholy, Ahmed, Gan, Cheng, and
  Liu}]{YenChunChen2020UNITERLU}
Yen-Chun Chen, Linjie Li, Licheng Yu, Ahmed~El Kholy, Faisal Ahmed, Zhe Gan,
  Yu~Cheng, and Jingjing Liu. 2020.
\newblock Uniter: Learning universal image-text representations.
\newblock \emph{european conference on computer vision}.

\bibitem[{Cho et~al.(2021)Cho, Lei, Tan, and Bansal}]{JaeminCho2021UnifyingVT}
Jaemin Cho, Jie Lei, Hao Tan, and Mohit Bansal. 2021.
\newblock \href {http://proceedings.mlr.press/v139/cho21a.html} {Unifying
  vision-and-language tasks via text generation}.
\newblock In \emph{Proceedings of the 38th International Conference on Machine
  Learning, {ICML} 2021, 18-24 July 2021, Virtual Event}, volume 139 of
  \emph{Proceedings of Machine Learning Research}, pages 1931--1942. {PMLR}.

\bibitem[{Cornia et~al.(2020)Cornia, Stefanini, Baraldi, and
  Cucchiara}]{MarcellaCornia2019MeshedMemoryTF}
Marcella Cornia, Matteo Stefanini, Lorenzo Baraldi, and Rita Cucchiara. 2020.
\newblock \href {https://doi.org/10.1109/CVPR42600.2020.01059} {Meshed-memory
  transformer for image captioning}.
\newblock In \emph{2020 {IEEE/CVF} Conference on Computer Vision and Pattern
  Recognition, {CVPR} 2020, Seattle, WA, USA, June 13-19, 2020}, pages
  10575--10584. {IEEE}.

\bibitem[{Das et~al.(2017)Das, Kottur, Gupta, Singh, Yadav, Moura, Parikh, and
  Batra}]{AbhishekDas2017VisualD}
Abhishek Das, Satwik Kottur, Khushi Gupta, Avi Singh, Deshraj Yadav, Jos{\'{e}}
  M.~F. Moura, Devi Parikh, and Dhruv Batra. 2017.
\newblock \href {https://doi.org/10.1109/CVPR.2017.121} {Visual dialog}.
\newblock In \emph{2017 {IEEE} Conference on Computer Vision and Pattern
  Recognition, {CVPR} 2017, Honolulu, HI, USA, July 21-26, 2017}, pages
  1080--1089. {IEEE} Computer Society.

\bibitem[{de~Vries et~al.(2017)de~Vries, Strub, Chandar, Pietquin, Larochelle,
  and Courville}]{HarmdeVries2016GuessWhatVO}
Harm de~Vries, Florian Strub, Sarath Chandar, Olivier Pietquin, Hugo
  Larochelle, and Aaron~C. Courville. 2017.
\newblock \href {https://doi.org/10.1109/CVPR.2017.475} {Guesswhat?! visual
  object discovery through multi-modal dialogue}.
\newblock In \emph{2017 {IEEE} Conference on Computer Vision and Pattern
  Recognition, {CVPR} 2017, Honolulu, HI, USA, July 21-26, 2017}, pages
  4466--4475. {IEEE} Computer Society.

\bibitem[{Deng et~al.(2018)Deng, Wu, Wu, Hu, Lyu, and
  Tan}]{ChaoruiDeng2018VisualGV}
Chaorui Deng, Qi~Wu, Qingyao Wu, Fuyuan Hu, Fan Lyu, and Mingkui Tan. 2018.
\newblock \href {https://doi.org/10.1109/CVPR.2018.00808} {Visual grounding via
  accumulated attention}.
\newblock In \emph{2018 {IEEE} Conference on Computer Vision and Pattern
  Recognition, {CVPR} 2018, Salt Lake City, UT, USA, June 18-22, 2018}, pages
  7746--7755. {IEEE} Computer Society.

\bibitem[{Devlin et~al.(2019)Devlin, Chang, Lee, and
  Toutanova}]{JacobDevlin2018BERTPO}
Jacob Devlin, Ming-Wei Chang, Kenton Lee, and Kristina Toutanova. 2019.
\newblock \href {https://doi.org/10.18653/v1/N19-1423} {{BERT}: Pre-training of
  deep bidirectional transformers for language understanding}.
\newblock In \emph{Proceedings of the 2019 Conference of the North {A}merican
  Chapter of the Association for Computational Linguistics: Human Language
  Technologies, Volume 1 (Long and Short Papers)}, pages 4171--4186,
  Minneapolis, Minnesota. Association for Computational Linguistics.

\bibitem[{Huang et~al.(2019)Huang, Wang, Chen, and
  Wei}]{LunHuang2019AttentionOA}
Lun Huang, Wenmin Wang, Jie Chen, and Xiaoyong Wei. 2019.
\newblock \href {https://doi.org/10.1109/ICCV.2019.00473} {Attention on
  attention for image captioning}.
\newblock In \emph{2019 {IEEE/CVF} International Conference on Computer Vision,
  {ICCV} 2019, Seoul, Korea (South), October 27 - November 2, 2019}, pages
  4633--4642. {IEEE}.

\bibitem[{Huang et~al.(2021)Huang, Zeng, Huang, Liu, Fu, and
  Fu}]{ZhichengHuang2021SeeingOO}
Zhicheng Huang, Zhaoyang Zeng, Yupan Huang, Bei Liu, Dongmei Fu, and Jianlong
  Fu. 2021.
\newblock Seeing out of the box: End-to-end pre-training for vision-language
  representation learning.
\newblock \emph{computer vision and pattern recognition}.

\bibitem[{Jin et~al.(2022)Jin, Cheng, Shen, Chen, and Ren}]{jin-etal-2022-good}
Woojeong Jin, Yu~Cheng, Yelong Shen, Weizhu Chen, and Xiang Ren. 2022.
\newblock \href {https://doi.org/10.18653/v1/2022.acl-long.197} {A good prompt
  is worth millions of parameters: Low-resource prompt-based learning for
  vision-language models}.
\newblock In \emph{Proceedings of the 60th Annual Meeting of the Association
  for Computational Linguistics (Volume 1: Long Papers)}, pages 2763--2775,
  Dublin, Ireland. Association for Computational Linguistics.

\bibitem[{Johnson et~al.(2016)Johnson, Karpathy, and
  Fei{-}Fei}]{JustinJohnson2015DenseCapFC}
Justin Johnson, Andrej Karpathy, and Li~Fei{-}Fei. 2016.
\newblock \href {https://doi.org/10.1109/CVPR.2016.494} {Densecap: Fully
  convolutional localization networks for dense captioning}.
\newblock In \emph{2016 {IEEE} Conference on Computer Vision and Pattern
  Recognition, {CVPR} 2016, Las Vegas, NV, USA, June 27-30, 2016}, pages
  4565--4574. {IEEE} Computer Society.

\bibitem[{Kamath et~al.(2021)Kamath, Singh, LeCun, Misra, Synnaeve, and
  Carion}]{AishwaryaKamath2021MDETRM}
Aishwarya Kamath, Mannat Singh, Yann LeCun, Ishan Misra, Gabriel Synnaeve, and
  Nicolas Carion. 2021.
\newblock Mdetr -- modulated detection for end-to-end multi-modal
  understanding.
\newblock \emph{international conference on computer vision}.

\bibitem[{Kazemzadeh et~al.(2014)Kazemzadeh, Ordonez, Matten, and
  Berg}]{SaharKazemzadeh2014ReferItGameRT}
Sahar Kazemzadeh, Vicente Ordonez, Mark Matten, and Tamara Berg. 2014.
\newblock \href {https://doi.org/10.3115/v1/D14-1086} {{R}efer{I}t{G}ame:
  Referring to objects in photographs of natural scenes}.
\newblock In \emph{Proceedings of the 2014 Conference on Empirical Methods in
  Natural Language Processing ({EMNLP})}, pages 787--798, Doha, Qatar.
  Association for Computational Linguistics.

\bibitem[{Kim et~al.(2020)Kim, Ko, and Wu}]{JungjunKim2020CoNANAC}
Jungjun Kim, Hanbin Ko, and Jialin Wu. 2020.
\newblock \href {https://doi.org/10.18653/v1/2020.coling-main.177} {{C}o{NAN}:
  A complementary neighboring-based attention network for referring expression
  generation}.
\newblock In \emph{Proceedings of the 28th International Conference on
  Computational Linguistics}, pages 1952--1962, Barcelona, Spain (Online).
  International Committee on Computational Linguistics.

\bibitem[{Kim et~al.(2021)Kim, Son, and Kim}]{WonjaeKim2021ViLTVT}
Wonjae Kim, Bokyung Son, and Ildoo Kim. 2021.
\newblock \href {http://proceedings.mlr.press/v139/kim21k.html} {Vilt:
  Vision-and-language transformer without convolution or region supervision}.
\newblock In \emph{Proceedings of the 38th International Conference on Machine
  Learning, {ICML} 2021, 18-24 July 2021, Virtual Event}, volume 139 of
  \emph{Proceedings of Machine Learning Research}, pages 5583--5594. {PMLR}.

\bibitem[{Krishna et~al.(2017)Krishna, Zhu, Groth, Johnson, Hata, Kravitz,
  Chen, Kalantidis, Li, Shamma, Bernstein, and
  Fei-Fei}]{RanjayKrishna2017VisualGC}
Ranjay Krishna, Yuke Zhu, Oliver Groth, Justin Johnson, Kenji Hata, Joshua
  Kravitz, Stephanie Chen, Yannis Kalantidis, Li-Jia Li, David~A. Shamma,
  Michael~S. Bernstein, and Li~Fei-Fei. 2017.
\newblock Visual genome: Connecting language and vision using crowdsourced
  dense image annotations.
\newblock \emph{International Journal of Computer Vision}.

\bibitem[{Lavie and Denkowski(2009)}]{AlonLavie2009TheMM}
Alon Lavie and Michael Denkowski. 2009.
\newblock The meteor metric for automatic evaluation of machine translation.
\newblock \emph{Machine Translation}.

\bibitem[{Li et~al.(2021{\natexlab{a}})Li, Selvaraju, Gotmare, Joty, Xiong, and
  Hoi}]{JunnanLi2021AlignBF}
Junnan Li, Ramprasaath~R. Selvaraju, Akhilesh Gotmare, Shafiq Joty, Caiming
  Xiong, and Steven C.~H. Hoi. 2021{\natexlab{a}}.
\newblock Align before fuse: Vision and language representation learning with
  momentum distillation.
\newblock \emph{neural information processing systems}.

\bibitem[{Li and Sigal(2021)}]{MuchenLi2021ReferringTA}
Muchen Li and Leonid Sigal. 2021.
\newblock Referring transformer: A one-step approach to multi-task visual
  grounding.
\newblock \emph{neural information processing systems}.

\bibitem[{Li et~al.(2021{\natexlab{b}})Li, Gao, Niu, Xiao, Liu, Liu, Wu, and
  Wang}]{WeiLi2021UNIMOTU}
Wei Li, Can Gao, Guocheng Niu, Xinyan Xiao, Hao Liu, Jiachen Liu, Hua Wu, and
  Haifeng Wang. 2021{\natexlab{b}}.
\newblock \href {https://doi.org/10.18653/v1/2021.acl-long.202} {{UNIMO}:
  Towards unified-modal understanding and generation via cross-modal
  contrastive learning}.
\newblock In \emph{Proceedings of the 59th Annual Meeting of the Association
  for Computational Linguistics and the 11th International Joint Conference on
  Natural Language Processing (Volume 1: Long Papers)}, pages 2592--2607,
  Online. Association for Computational Linguistics.

\bibitem[{Li et~al.(2020)Li, Yin, Li, Zhang, Hu, Zhang, Wang, Hu, Dong, Wei,
  Choi, and Gao}]{DBLP:journals/corr/abs-2004-06165}
Xiujun Li, Xi~Yin, Chunyuan Li, Pengchuan Zhang, Xiaowei Hu, Lei Zhang, Lijuan
  Wang, Houdong Hu, Li~Dong, Furu Wei, Yejin Choi, and Jianfeng Gao. 2020.
\newblock Oscar: Object-semantics aligned pre-training for vision-language
  tasks.
\newblock \emph{european conference on computer vision}.

\bibitem[{Lin et~al.(2014)Lin, Maire, Belongie, Hays, Perona, Ramanan,
  Doll{\'a}r, and Zitnick}]{TsungYiLin2014MicrosoftCC}
Tsung-Yi Lin, Michael Maire, Serge Belongie, James Hays, Pietro Perona, Deva
  Ramanan, Piotr Doll{\'a}r, and C.~Lawrence Zitnick. 2014.
\newblock Microsoft coco: Common objects in context.
\newblock \emph{european conference on computer vision}.

\bibitem[{Liu et~al.(2020)Liu, Wang, Wang, and
  Yang}]{JingyuLiu2020AttributeGuidedAF}
Jingyu Liu, Wei Wang, Liang Wang, and Ming-Hsuan Yang. 2020.
\newblock Attribute-guided attention for referring expression generation and
  comprehension.
\newblock \emph{IEEE Transactions on Image Processing}.

\bibitem[{Liu et~al.(2021)Liu, Feng, and Wang}]{TongtongLiu2021MultistagePO}
Tongtong Liu, Fangxiang Feng, and Xiaojie Wang. 2021.
\newblock \href {https://doi.org/10.18653/v1/2021.acl-long.199} {Multi-stage
  pre-training over simplified multimodal pre-training models}.
\newblock In \emph{Proceedings of the 59th Annual Meeting of the Association
  for Computational Linguistics and the 11th International Joint Conference on
  Natural Language Processing (Volume 1: Long Papers)}, pages 2556--2565,
  Online. Association for Computational Linguistics.

\bibitem[{Lu et~al.(2019)Lu, Batra, Parikh, and Lee}]{JiasenLu2019ViLBERTPT}
Jiasen Lu, Dhruv Batra, Devi Parikh, and Stefan Lee. 2019.
\newblock \href
  {https://proceedings.neurips.cc/paper/2019/hash/c74d97b01eae257e44aa9d5bade97baf-Abstract.html}
  {Vilbert: Pretraining task-agnostic visiolinguistic representations for
  vision-and-language tasks}.
\newblock In \emph{Advances in Neural Information Processing Systems 32: Annual
  Conference on Neural Information Processing Systems 2019, NeurIPS 2019,
  December 8-14, 2019, Vancouver, BC, Canada}, pages 13--23.

\bibitem[{Luo and Shakhnarovich(2017)}]{RuotianLuo2017ComprehensionguidedRE}
Ruotian Luo and Gregory Shakhnarovich. 2017.
\newblock \href {https://doi.org/10.1109/CVPR.2017.333} {Comprehension-guided
  referring expressions}.
\newblock In \emph{2017 {IEEE} Conference on Computer Vision and Pattern
  Recognition, {CVPR} 2017, Honolulu, HI, USA, July 21-26, 2017}, pages
  3125--3134. {IEEE} Computer Society.

\bibitem[{Mao et~al.(2016)Mao, Huang, Toshev, Camburu, Yuille, and
  Murphy}]{JunhuaMao2015GenerationAC}
Junhua Mao, Jonathan Huang, Alexander Toshev, Oana Camburu, Alan~L. Yuille, and
  Kevin Murphy. 2016.
\newblock \href {https://doi.org/10.1109/CVPR.2016.9} {Generation and
  comprehension of unambiguous object descriptions}.
\newblock In \emph{2016 {IEEE} Conference on Computer Vision and Pattern
  Recognition, {CVPR} 2016, Las Vegas, NV, USA, June 27-30, 2016}, pages
  11--20. {IEEE} Computer Society.

\bibitem[{Plummer et~al.(2015)Plummer, Wang, Cervantes, Caicedo, Hockenmaier,
  and Lazebnik}]{BryanAPlummer2015Flickr30kEC}
Bryan~A. Plummer, Liwei Wang, Chris~M. Cervantes, Juan~C. Caicedo, Julia
  Hockenmaier, and Svetlana Lazebnik. 2015.
\newblock \href {https://doi.org/10.1109/ICCV.2015.303} {Flickr30k entities:
  Collecting region-to-phrase correspondences for richer image-to-sentence
  models}.
\newblock In \emph{2015 {IEEE} International Conference on Computer Vision,
  {ICCV} 2015, Santiago, Chile, December 7-13, 2015}, pages 2641--2649. {IEEE}
  Computer Society.

\bibitem[{Radford et~al.(2021)Radford, Kim, Hallacy, Ramesh, Goh, Agarwal,
  Sastry, Askell, Mishkin, Clark, Krueger, and
  Sutskever}]{AlecRadford2021LearningTV}
Alec Radford, Jong~Wook Kim, Chris Hallacy, Aditya Ramesh, Gabriel Goh,
  Sandhini Agarwal, Girish Sastry, Amanda Askell, Pamela Mishkin, Jack Clark,
  Gretchen Krueger, and Ilya Sutskever. 2021.
\newblock \href {http://proceedings.mlr.press/v139/radford21a.html} {Learning
  transferable visual models from natural language supervision}.
\newblock In \emph{Proceedings of the 38th International Conference on Machine
  Learning, {ICML} 2021, 18-24 July 2021, Virtual Event}, volume 139 of
  \emph{Proceedings of Machine Learning Research}, pages 8748--8763. {PMLR}.

\bibitem[{Ren et~al.(2015)Ren, He, Girshick, and Sun}]{ShaoqingRen2015FasterRT}
Shaoqing Ren, Kaiming He, Ross~B. Girshick, and Jian Sun. 2015.
\newblock \href
  {https://proceedings.neurips.cc/paper/2015/hash/14bfa6bb14875e45bba028a21ed38046-Abstract.html}
  {Faster {R-CNN:} towards real-time object detection with region proposal
  networks}.
\newblock In \emph{Advances in Neural Information Processing Systems 28: Annual
  Conference on Neural Information Processing Systems 2015, December 7-12,
  2015, Montreal, Quebec, Canada}, pages 91--99.

\bibitem[{Rezatofighi et~al.(2019)Rezatofighi, Tsoi, Gwak, Sadeghian, Reid, and
  Savarese}]{HamidRezatofighi2019GeneralizedIO}
Hamid Rezatofighi, Nathan Tsoi, JunYoung Gwak, Amir Sadeghian, Ian~D. Reid, and
  Silvio Savarese. 2019.
\newblock \href {https://doi.org/10.1109/CVPR.2019.00075} {Generalized
  intersection over union: {A} metric and a loss for bounding box regression}.
\newblock In \emph{{IEEE} Conference on Computer Vision and Pattern
  Recognition, {CVPR} 2019, Long Beach, CA, USA, June 16-20, 2019}, pages
  658--666. Computer Vision Foundation / {IEEE}.

\bibitem[{Rohrbach et~al.(2015)Rohrbach, Rohrbach, Hu, Darrell, and
  Schiele}]{AnnaRohrbach2015GroundingOT}
Anna Rohrbach, Marcus Rohrbach, Ronghang Hu, Trevor Darrell, and Bernt Schiele.
  2015.
\newblock Grounding of textual phrases in images by reconstruction.
\newblock \emph{european conference on computer vision}.

\bibitem[{Su et~al.(2020)Su, Zhu, Cao, Li, Lu, Wei, and
  Dai}]{WeijieSu2020VLBERTPO}
Weijie Su, Xizhou Zhu, Yue Cao, Bin Li, Lewei Lu, Furu Wei, and Jifeng Dai.
  2020.
\newblock \href {https://openreview.net/forum?id=SygXPaEYvH} {{VL-BERT:}
  pre-training of generic visual-linguistic representations}.
\newblock In \emph{8th International Conference on Learning Representations,
  {ICLR} 2020, Addis Ababa, Ethiopia, April 26-30, 2020}. OpenReview.net.

\bibitem[{Sun et~al.(2022)Sun, Suo, Wang, Zhang, and Wu}]{9699024}
Mengyang Sun, Wei Suo, Peng Wang, Yanning Zhang, and Qi~Wu. 2022.
\newblock \href {https://doi.org/10.1109/TMM.2022.3147385} {A proposal-free
  one-stage framework for referring expression comprehension and generation via
  dense cross-attention}.
\newblock \emph{IEEE Transactions on Multimedia}, pages 1--1.

\bibitem[{Tan and Bansal(2019)}]{HaoTan2019LXMERTLC}
Hao Tan and Mohit Bansal. 2019.
\newblock \href {https://doi.org/10.18653/v1/D19-1514} {{LXMERT}: Learning
  cross-modality encoder representations from transformers}.
\newblock In \emph{Proceedings of the 2019 Conference on Empirical Methods in
  Natural Language Processing and the 9th International Joint Conference on
  Natural Language Processing (EMNLP-IJCNLP)}, pages 5100--5111, Hong Kong,
  China. Association for Computational Linguistics.

\bibitem[{Tanaka et~al.(2019{\natexlab{a}})Tanaka, Itamochi, Narioka, Sato,
  Ushiku, and Harada}]{MikihiroTanaka2018GeneratingER}
Mikihiro Tanaka, Takayuki Itamochi, Kenichi Narioka, Ikuro Sato, Yoshitaka
  Ushiku, and Tatsuya Harada. 2019{\natexlab{a}}.
\newblock \href {https://doi.org/10.1109/ICCV.2019.00589} {Generating
  easy-to-understand referring expressions for target identifications}.
\newblock In \emph{2019 {IEEE/CVF} International Conference on Computer Vision,
  {ICCV} 2019, Seoul, Korea (South), October 27 - November 2, 2019}, pages
  5793--5802. {IEEE}.

\bibitem[{Tanaka et~al.(2019{\natexlab{b}})Tanaka, Itamochi, Narioka, Sato,
  Ushiku, and Harada}]{Tanaka_2019_ICCV}
Mikihiro Tanaka, Takayuki Itamochi, Kenichi Narioka, Ikuro Sato, Yoshitaka
  Ushiku, and Tatsuya Harada. 2019{\natexlab{b}}.
\newblock Generating easy-to-understand referring expressions for target
  identifications.
\newblock In \emph{Proceedings of the IEEE/CVF International Conference on
  Computer Vision (ICCV)}.

\bibitem[{Tsimpoukelli et~al.(2021)Tsimpoukelli, Menick, Cabi, Eslami, Vinyals,
  and Hill}]{MariaTsimpoukelli2021MultimodalFL}
Maria Tsimpoukelli, Jacob Menick, Serkan Cabi, S.~M.~Ali Eslami, Oriol Vinyals,
  and Felix Hill. 2021.
\newblock Multimodal few-shot learning with frozen language models.
\newblock \emph{neural information processing systems}.

\bibitem[{Vedantam et~al.(2015)Vedantam, Zitnick, and
  Parikh}]{RamakrishnaVedantam2014CIDErCI}
Ramakrishna Vedantam, C.~Lawrence Zitnick, and Devi Parikh. 2015.
\newblock \href {https://doi.org/10.1109/CVPR.2015.7299087} {Cider:
  Consensus-based image description evaluation}.
\newblock In \emph{{IEEE} Conference on Computer Vision and Pattern
  Recognition, {CVPR} 2015, Boston, MA, USA, June 7-12, 2015}, pages
  4566--4575. {IEEE} Computer Society.

\bibitem[{Wang et~al.(2019)Wang, Wu, Cao, Shen, Gao, and van~den
  Hengel}]{DBLP:journals/corr/abs-1812-04794}
Peng Wang, Qi~Wu, Jiewei Cao, Chunhua Shen, Lianli Gao, and Anton van~den
  Hengel. 2019.
\newblock \href {https://doi.org/10.1109/CVPR.2019.00206} {Neighbourhood watch:
  Referring expression comprehension via language-guided graph attention
  networks}.
\newblock In \emph{{IEEE} Conference on Computer Vision and Pattern
  Recognition, {CVPR} 2019, Long Beach, CA, USA, June 16-20, 2019}, pages
  1960--1968. Computer Vision Foundation / {IEEE}.

\bibitem[{Wang et~al.(2022)Wang, Yang, Men, Lin, Bai, Li, Ma, Zhou, Zhou, Yang,
  and Zhou<ericzhou}]{PengWang2022UNIFYINGAT}
Peng Wang, An~Yang, Rui Men, Junyang Lin, Shuai Bai, Zhikang Li, Jianxin Ma,
  Chang Zhou, Jingren Zhou, Hongxia Yang, and Chang Zhou<ericzhou. 2022.
\newblock Unifying architectures, tasks, and modalities through a simple
  sequence-to-sequence learning framework.

\bibitem[{Wang et~al.(20{\natexlab{a}})Wang, Bao, Dong, and
  Wei}]{WenhuiWang2021VLMoUV}
Wenhui Wang, Hangbo Bao, Li~Dong, and Furu Wei. 20{\natexlab{a}}.
\newblock \href {https://arxiv.org/abs/} {Vlmo: Unified vision-language
  pre-training with mixture-of-modality-experts}.
\newblock \emph{ArXiv preprint}, abs/.

\bibitem[{Wang et~al.(20{\natexlab{b}})Wang, Yu, Yu, Dai, Tsvetkov, and
  Cao}]{ZiruiWang2021SimVLMSV}
Zirui Wang, Jiahui Yu, Adams~Wei Yu, Zihang Dai, Yulia Tsvetkov, and Yuan Cao.
  20{\natexlab{b}}.
\newblock \href {https://arxiv.org/abs/} {Simvlm: Simple visual language model
  pretraining with weak supervision}.
\newblock \emph{ArXiv preprint}, abs/.

\bibitem[{Wu et~al.(2020)Wu, Lin, Cohen, Bui, and
  Maji}]{ChenyunWu2020PhraseCutLI}
Chenyun Wu, Zhe Lin, Scott Cohen, Trung Bui, and Subhransu Maji. 2020.
\newblock \href {https://doi.org/10.1109/CVPR42600.2020.01023} {Phrasecut:
  Language-based image segmentation in the wild}.
\newblock In \emph{2020 {IEEE/CVF} Conference on Computer Vision and Pattern
  Recognition, {CVPR} 2020, Seattle, WA, USA, June 13-19, 2020}, pages
  10213--10222. {IEEE}.

\bibitem[{Wu et~al.(20)Wu, Schuster, Chen, Le, Norouzi, Macherey, Krikun, Cao,
  Gao, Macherey, Klingner, Shah, Johnson, Liu, Łukasz Kaiser, Gouws, Kato,
  Kudo, Kazawa, Stevens, Kurian, Patil, Wang, Young, Smith, Riesa, Rudnick,
  Vinyals, Corrado, Hughes, and Dean}]{YonghuiWu2016GooglesNM}
Yonghui Wu, Mike Schuster, Zhifeng Chen, Quoc~V. Le, Mohammad Norouzi, Wolfgang
  Macherey, Maxim Krikun, Yuan Cao, Qin Gao, Klaus Macherey, Jeff Klingner,
  Apurva Shah, Melvin Johnson, Xiaobing Liu, Łukasz Kaiser, Stephan Gouws,
  Yoshikiyo Kato, Taku Kudo, Hideto Kazawa, Keith Stevens, George Kurian,
  Nishant Patil, Wei Wang, Cliff Young, Jason~A. Smith, Jason Riesa, Alex
  Rudnick, Oriol Vinyals, Greg~S. Corrado, Macduff Hughes, and Jeffrey Dean.
  20.
\newblock \href {https://arxiv.org/abs/} {Google's neural machine translation
  system: Bridging the gap between human and machine translation}.
\newblock \emph{ArXiv preprint}, abs/.

\bibitem[{Yang et~al.(2020)Yang, Chen, Wang, and
  Luo}]{ZhengyuanYang2020ImprovingOV}
Zhengyuan Yang, Tianlang Chen, Liwei Wang, and Jiebo Luo. 2020.
\newblock Improving one-stage visual grounding by recursive sub-query
  construction.
\newblock \emph{european conference on computer vision}.

\bibitem[{Yang et~al.(2019)Yang, Gong, Wang, Huang, Yu, and
  Luo}]{ZhengyuanYang2019AFA}
Zhengyuan Yang, Boqing Gong, Liwei Wang, Wenbing Huang, Dong Yu, and Jiebo Luo.
  2019.
\newblock \href {https://doi.org/10.1109/ICCV.2019.00478} {A fast and accurate
  one-stage approach to visual grounding}.
\newblock In \emph{2019 {IEEE/CVF} International Conference on Computer Vision,
  {ICCV} 2019, Seoul, Korea (South), October 27 - November 2, 2019}, pages
  4682--4692. {IEEE}.

\bibitem[{Yu et~al.(2020)Yu, Tang, Yin, Sun, Tian, Wu, and
  Wang}]{FeiYu2020ERNIEViLKE}
Fei Yu, Jiji Tang, Weichong Yin, Yu~Sun, Hao Tian, Hua Wu, and Haifeng Wang.
  2020.
\newblock Ernie-vil: Knowledge enhanced vision-language representations through
  scene graphs.
\newblock \emph{national conference on artificial intelligence}.

\bibitem[{Yu et~al.(2018)Yu, Lin, Shen, Yang, Lu, Bansal, and
  Berg}]{LichengYu2018MAttNetMA}
Licheng Yu, Zhe Lin, Xiaohui Shen, Jimei Yang, Xin Lu, Mohit Bansal, and
  Tamara~L. Berg. 2018.
\newblock \href {https://doi.org/10.1109/CVPR.2018.00142} {Mattnet: Modular
  attention network for referring expression comprehension}.
\newblock In \emph{2018 {IEEE} Conference on Computer Vision and Pattern
  Recognition, {CVPR} 2018, Salt Lake City, UT, USA, June 18-22, 2018}, pages
  1307--1315. {IEEE} Computer Society.

\bibitem[{Yu et~al.(2016)Yu, Poirson, Yang, Berg, and
  Berg}]{LichengYu2016ModelingCI}
Licheng Yu, Patrick Poirson, Shan Yang, Alexander~C. Berg, and Tamara~L. Berg.
  2016.
\newblock Modeling context in referring expressions.
\newblock \emph{european conference on computer vision}.

\bibitem[{Yu et~al.(2017)Yu, Tan, Bansal, and Berg}]{LichengYu2016AJS}
Licheng Yu, Hao Tan, Mohit Bansal, and Tamara~L. Berg. 2017.
\newblock \href {https://doi.org/10.1109/CVPR.2017.375} {A joint
  speaker-listener-reinforcer model for referring expressions}.
\newblock In \emph{2017 {IEEE} Conference on Computer Vision and Pattern
  Recognition, {CVPR} 2017, Honolulu, HI, USA, July 21-26, 2017}, pages
  3521--3529. {IEEE} Computer Society.

\bibitem[{Zeng et~al.(2021)Zeng, Zhang, and Li}]{YanZeng2021MultiGrainedVL}
Yan Zeng, Xinsong Zhang, and Hang Li. 2021.
\newblock \href {https://arxiv.org/abs/} {Multi-grained vision language
  pre-training: Aligning texts with visual concepts}.
\newblock \emph{ArXiv preprint}, abs/.

\bibitem[{Zhou et~al.(2020)Zhou, Palangi, Zhang, Hu, Corso, and
  Gao}]{LuoweiZhou2020UnifiedVP}
Luowei Zhou, Hamid Palangi, Lei Zhang, Houdong Hu, Jason~J. Corso, and Jianfeng
  Gao. 2020.
\newblock \href {https://aaai.org/ojs/index.php/AAAI/article/view/7005}
  {Unified vision-language pre-training for image captioning and {VQA}}.
\newblock In \emph{The Thirty-Fourth {AAAI} Conference on Artificial
  Intelligence, {AAAI} 2020, The Thirty-Second Innovative Applications of
  Artificial Intelligence Conference, {IAAI} 2020, The Tenth {AAAI} Symposium
  on Educational Advances in Artificial Intelligence, {EAAI} 2020, New York,
  NY, USA, February 7-12, 2020}, pages 13041--13049. {AAAI} Press.

\end{thebibliography}
\bibliographystyle{acl_natbib}

\appendix
\begin{figure*}[!ht] 
\centering 
\includegraphics[width=1\textwidth]{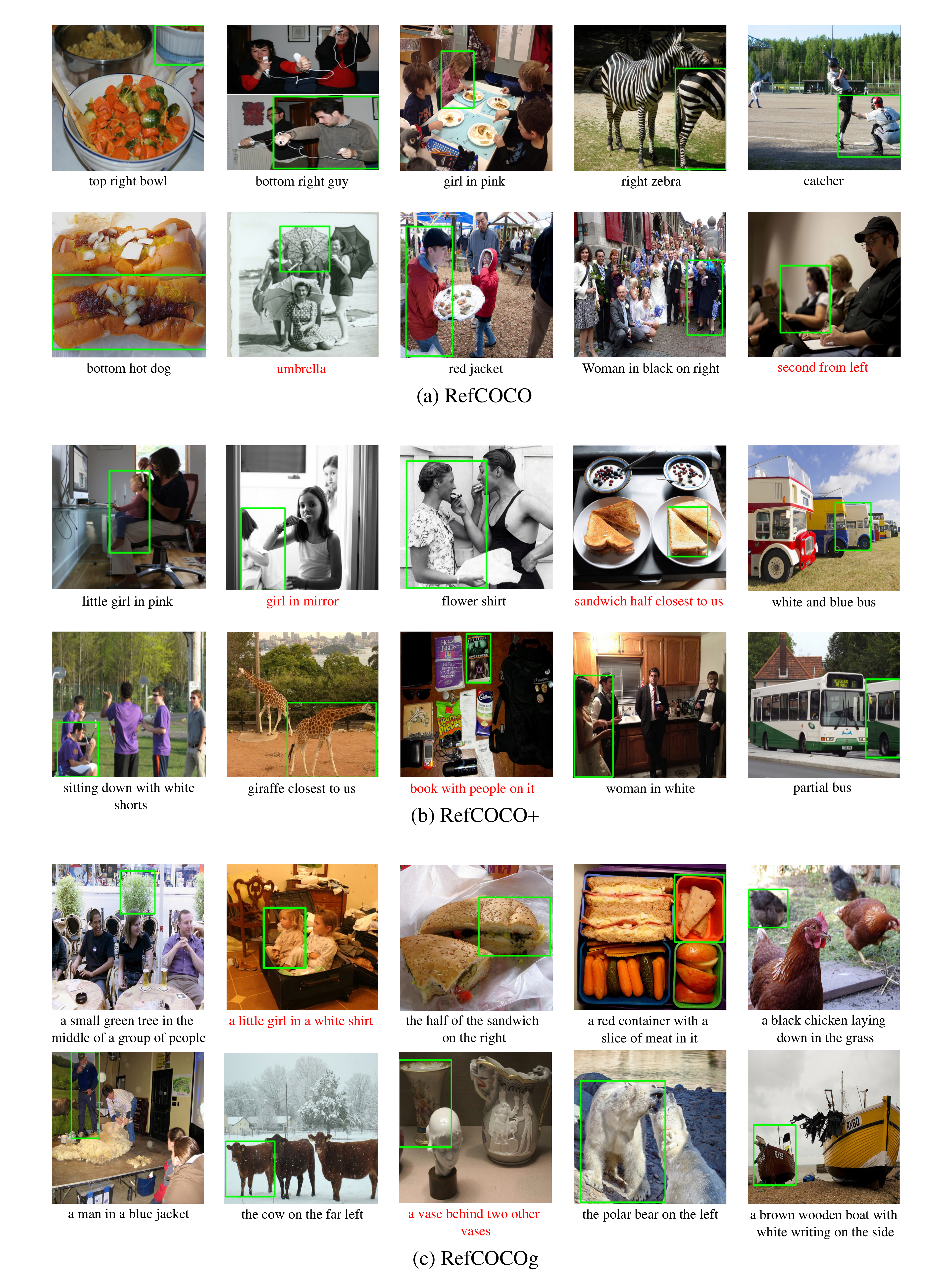} 
\vspace{-0.4cm}
\caption{Uncurated examples of the generated text in REG. The inaccurate or ambiguous text is marked in red.}
\vspace{-0.2cm}
\label{examples_reg}
\end{figure*}

\begin{figure*}[!ht] 
\centering 
\includegraphics[width=1\textwidth]{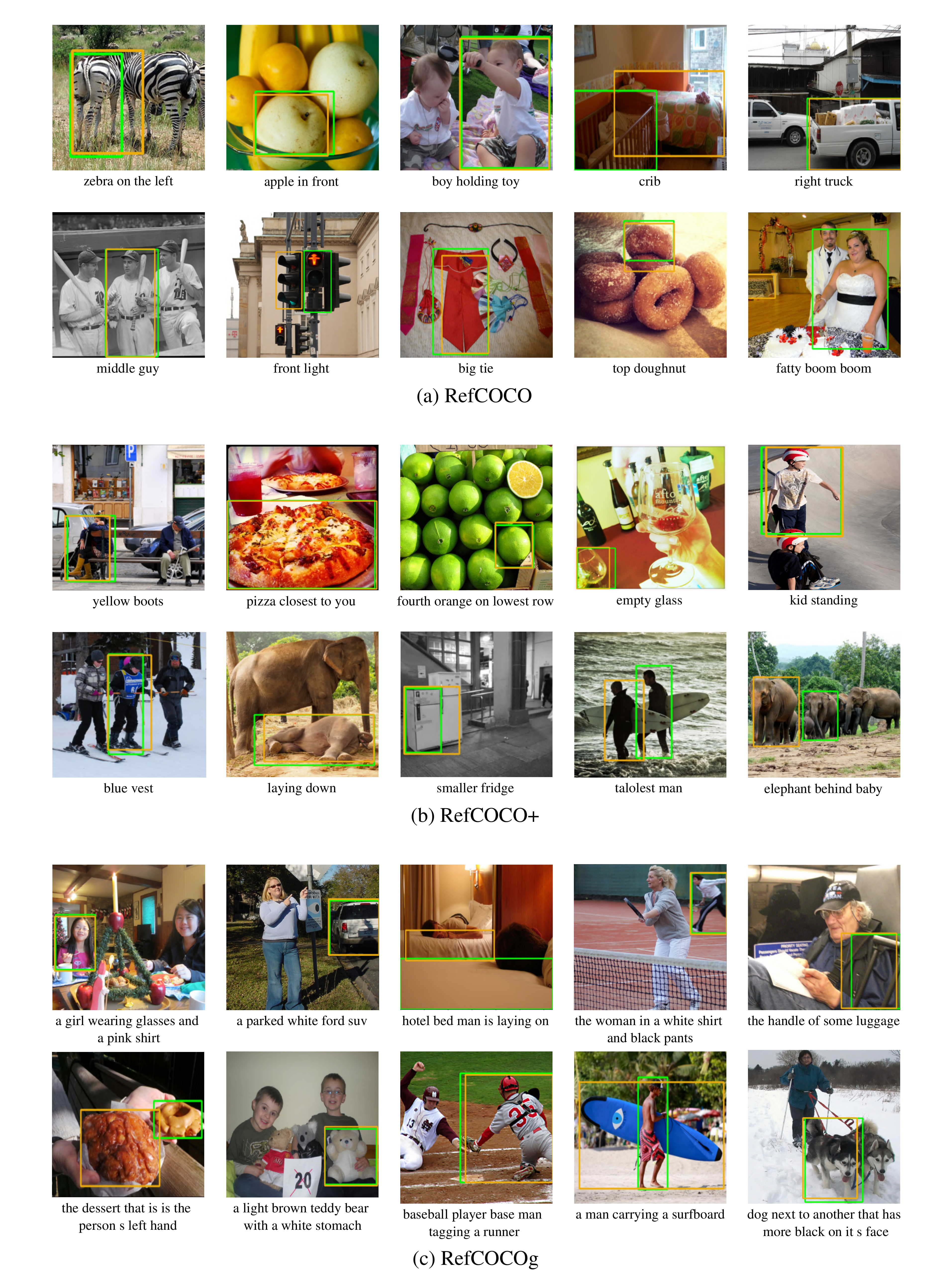} 
\vspace{-0.4cm}
\caption{Uncurated examples of the predicted bounding box in REC. The green and orange boxes indicate the ground truth and prediction, respectively.}
\vspace{-0.2cm}
\label{examples_rec}
\end{figure*}

\section{Examples on REG and REC} \label{section::examples}
We give more uncurated examples on REG and REC in Fig. \ref{examples_reg} and \ref{examples_rec}, respectively.

\end{document}